\pdfoutput=1
\documentclass[11pt]{article}

\usepackage[final]{acl}

\usepackage{times}
\usepackage{latexsym}

\usepackage[T1]{fontenc}
\usepackage[utf8]{inputenc}

\usepackage{microtype}

\usepackage{inconsolata}

\usepackage{graphicx}

\usepackage{hyperref}
\usepackage{url}
\usepackage{bm}
\usepackage{multirow}
\usepackage{array}
\usepackage{booktabs}
\usepackage{amsmath}
\usepackage{cleveref}
\usepackage{pifont}
\usepackage{enumitem}

\newcommand{\red}[1]{{\color{red}{#1}}}

\newcommand{\gamenospace}{\texttt{GAME}}
\newcommand{\game}{\texttt{GAME} }

\title{Gradient-guided Attention Map Editing: Towards Efficient Contextual Hallucination Mitigation}

\author{
 \textbf{Yu Wang\textsuperscript{1\thanks{The work is done during the summer internship at Intuit.}}},
 \textbf{Kamalika Das\textsuperscript{2}},
 \textbf{Xiang Gao\textsuperscript{2}},
 \textbf{Wendi Cui\textsuperscript{2}},
 \textbf{Peng Li\textsuperscript{1}},
 \textbf{Jiaxin Zhang\textsuperscript{2}}
\\
 \textsuperscript{1}University of California, Santa Barbara, Santa Barbara, CA
\\
 \textsuperscript{2}Intuit AI Research, Mountain View, CA
\\
 \small{
    \{yu95, lip\}@ucsb.edu
 }
\\
 \small{
    \{Jiaxin\_Zhang, Xiang\_Gao, Wendi\_Cui, Kamalika\_Das\}@intuit.com
 }
}

\begin{document}
\maketitle
\begin{abstract}
In tasks like summarization and open-book question answering (QA), Large Language Models (LLMs) often encounter "contextual hallucination", where they produce irrelevant or incorrect responses despite having access to accurate source information. This typically occurs because these models tend to prioritize self-generated content over the input context, causing them to disregard pertinent details. To address this challenge, we introduce a novel method called "Guided Attention Map Editing" (\gamenospace), which dynamically adjusts attention maps to improve contextual relevance. During inference, \game employs a trained classifier to identify attention maps prone to inducing hallucinations and executes targeted interventions. These interventions, guided by gradient-informed ``edit directions'', strategically redistribute attention weights across various heads to effectively reduce hallucination. Comprehensive evaluations on challenging summarization and open-book QA tasks show that \game consistently reduces hallucinations across a variety of open-source models. Specifically, \game reduces hallucinations by {\bf 10\%} in the XSum summarization task while achieving a {\bf 7X} speed-up in computational efficiency compared to the state-of-the-art baselines.

\end{abstract}

\section{Introduction}

Large Language Models (LLMs) have demonstrated remarkable capabilities across various natural language processing tasks. Despite these advances, their practical deployment is often compromised by their propensity to produce hallucinated outputs \cite{huang2023survey,ji2023survey, tonmoy2024comprehensive}. Hallucinations arise from multiple sources, and significant research efforts have focused on detecting and mitigating them. For example, hallucinations due to outdated or incomplete knowledge in the training data can be addressed through techniques such as Retrieval-Augmented Generation (RAG) \cite{rag_survey} or knowledge injection \cite{ovadia2023fine, zhang2024synthetic}. Recent methods that intervene in the hidden representations\cite{iti, plug_and_play, dola} of language models during inference have shown substantial improvements in reducing hallucinations related to inherent confirmation bias and spurious correlations \cite{overshadow}.

This work mainly targets a particularly critical phenomenon known as ``contextual hallucination'' \cite{ainsworth2024reducing, lb_lens}, also referred to as ``openbook hallucination'' \cite{constructing_benchmark_hallucination}. This occurs when LLMs generate misleading or unrelated content despite having access to pertinent information in the input context. Contextual hallucination presents significant challenges, particularly in high-stakes domains such as finance and healthcare, where accurate retrieval of relevant information is crucial. Failures in these contexts can lead to severe consequences, underscoring the urgent need for effective mitigation strategies.

Contextual hallucination represents an inherent deficiency within LLMs that cannot be resolved merely by injecting new knowledge or providing additional information, which makes it a difficult challenge. It often arises from a poor correlation between the input context and the generated outputs. Previous efforts, such as those in \citet{contextual_aware_decoding} and \citet{mutual_alleviate}, have encouraged enhancing the mutual correlation between the input context and generation to reduce hallucinations. More recently, \citet{lb_lens} explores deep into the architecture of transformers, suggesting that contextual hallucinations stem from a suboptimal distribution of attention between context and self-generation. The studies also demonstrate that identifying problematic attention maps can be used to model and mitigate contextual hallucinations. However, while promising, these methods fail to actively intervene in the attention maps but rather select the most promising outputs from multiple random samples. These passive approaches, relying heavily on the inherent capabilities of LLMs, can be inefficient and may not consistently correct the root cause of hallucinations.

Inspired by these insights, we argue that directly modifying problematic attention maps shall be a more effective way to reduce contextual hallucination. This intervention encourages the model to focus more on pertinent content, enhancing relevance and coherence in the generated text. To assess this hypothesis, we initiated a behavioral study to evaluate the impact of attention editing on LLMs. We designed experiments to bias the model’s attention towards the context during inference, aiming to guide the model to prioritize contextually relevant information.

Our preliminary findings reveal that directing attention towards contextual elements can prompt the model to produce more grounded and contextually relevant content, thus minimizing instances of ungrounded generation. However, the study also underscores the importance of precise attention editing. Arbitrary modifications to attention maps can disrupt the natural functioning of LLMs and may inadvertently introduce additional errors or hallucinations. This highlights the critical need for targeted and carefully calibrated interventions in attention mechanisms to avoid unintended consequences.

To address these challenges, we propose ``Guided Attention Map Editing'' (\gamenospace), a method designed to perform precise interventions on attention maps to reduce contextual hallucination. \game employs a classifier trained to identify problematic attention maps that are likely to induce hallucinations during inference. Once such maps are detected, \game triggers an intervention to regenerate the corresponding outputs. This process involves using gradient information, referred to as ``edit direction'', to reallocate attention weights across different attention heads strategically. Importantly, the edit direction is tailored for each head, recognizing important findings in past literature on the diverse functionality of different attention heads \cite{attn_survey}. The proposed \game can effectively reduce the hallucination rate by $10\%$ of Llama2-7b on XSum \cite{xsum} dataset with negligible additional computation cost. The contributions of this work are threefold: 

\begin{itemize}[leftmargin=10pt] 
    \item We conduct a behavioral study to investigate the impact of editing attention maps in producing more contextually aware content and highlight the importance of editing direction.
    \item We propose \gamenospace, a novel approach leveraging gradient information to guide the editing of attention map precisely, reducing contextual hallucinations efficiently.
    \item We demonstrate \game on extensive experiments, including summarization and open-book QA tasks, showing superior performance compared to the state-of-the-art baseline methods. 
\end{itemize}

\section{Preliminaries}
\subsection{Attention Mechanism in Transformers}
\begin{figure}[h]
    \centering
    \includegraphics[width=1\linewidth]{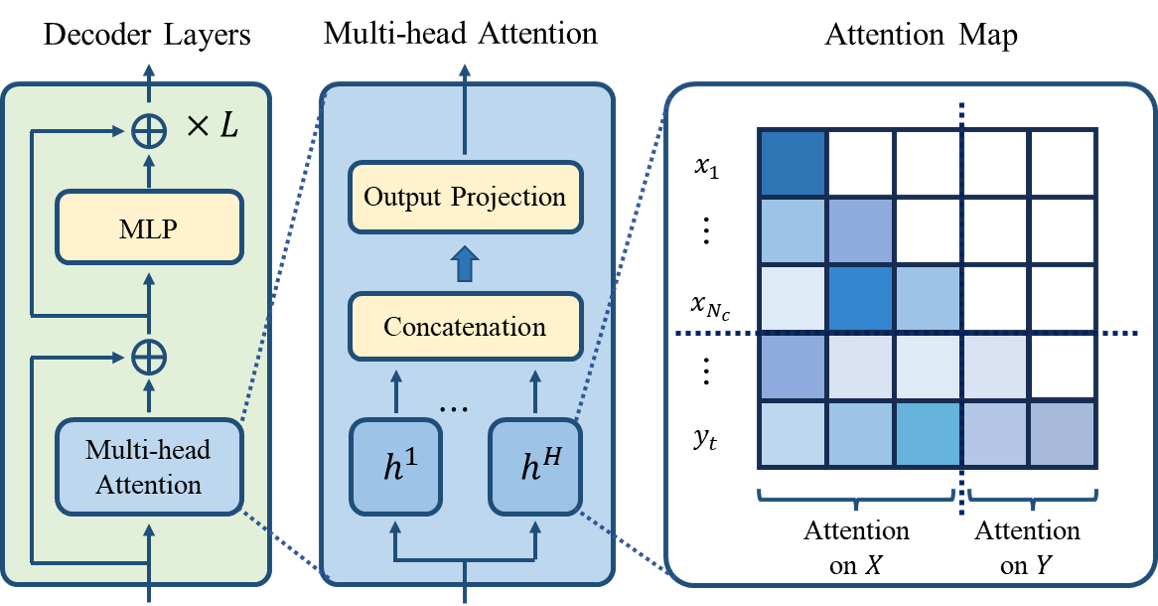}
    \caption{Illustration of a decoder-only Transformer featuring a multi-head attention mechanism. Each row in an attention map represents a weight vector that sums to one, reflecting the current token's relationship with preceding tokens. A deeper color indicates a higher attention weight.}
    \label{fig:attn}
\end{figure}

Most LLMs today utilize the Transformer architecture \cite{vaswani2017attention}, which predominantly features a decoder-only architecture and employs multi-head attention mechanism \cite{zheng2024attention}  to effectively handle the complex correlations among input tokens, as shown in      \Cref{fig:attn}.

We examine a Transformer model consisting of $ L $ decoder layers, each equipped with $ H $ attention heads, indexed by $ l $ and $ h $ for the layer and head, respectively. The model processes input with length $N$, which is a concatenation of the context $ X = [x_1, x_2, \ldots, x_{N_c}]^{N_c} $ and the preceding generated sequence $ Y = [y_1, y_2, \ldots, y_{t}]^{N_g} $, where $ N_c $ and $ N_g $ denote the lengths of the context and generated sequence, respectively. 

As the model predicts the next token, each attention head independently computes an attention map from the input. These maps are lower triangular matrices where each row consists of weights that sum to one, reflecting the relationship of the current token with preceding tokens. A higher weight within this matrix indicates a stronger correlation, suggesting that the model is more likely to generate tokens closely related to those with higher weights. Importantly, different attention heads are tailored to focus on various aspects of the input, allowing for a comprehensive, multi-faceted analysis of token relationships.

\subsection{Attention Feature for Detecting Contextual Hallucination}

Contextual hallucination occurs when LLMs generate outputs that do not exist or cannot be inferred from the provided context. Past literature \cite{lb_lens} has demonstrated that the model's lack of attention to the context during the generation process can be the cause and utilizing the attention map as a feature can effectively detect such contextual hallucination.
\begin{figure}
    \centering
    \includegraphics[width=0.97\linewidth]{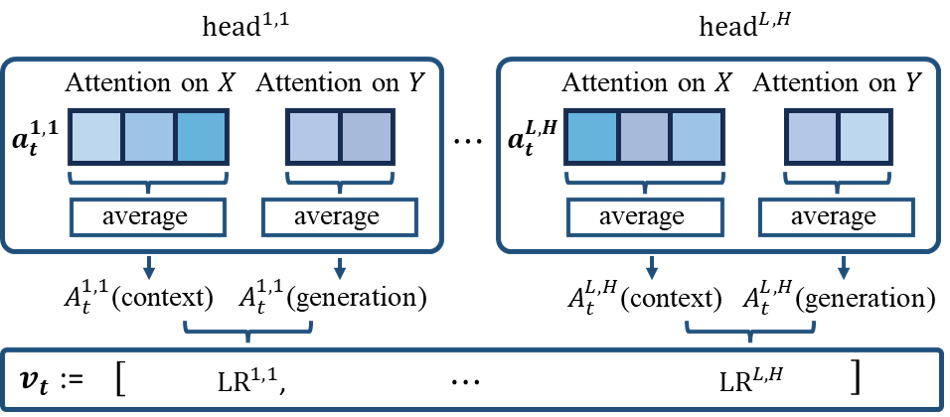}
    \caption{Derivation of the LR at the $t_{\text{th}}$ decoding step.}
    \label{fig:lr}
\end{figure}
Specifically, an attention map based feature named ``Lookback Ratio'' (LR) is proposed to model contextual hallucination. As shown in \Cref{fig:lr}, at the $t_{\text{th}}$ decoding step (to predict $y_{t+1}$), for the $h_{\text{th}}$ head in the $l_{\text{th}}$ layer, its LR is defined as: 
\begin{equation}
    \text{LR}^{l,h}_t = \frac{A^{l,h}_t(\text{context})}{A^{l,h}_t(\text{context}) + A^{l,h}_t(\text{generation})},
\end{equation}
where
\begin{equation}
\begin{aligned}
    &A^{l,h}_t(\text{context}) = \frac{1}{N_{c}} \sum_{i=1}^{N_{c}} a^{l,h}_{i},\\
    &A^{l,h}_t(\text{generation}) = \frac{1}{N_{g}} \sum_{j=N_{c}+1}^{N} a^{l,h}_{j},
\end{aligned}
\end{equation}
$\bm{a^{l,h}}$ denotes the post-Softmax attention weights of the head. Across all heads, the lookback ratio vector is defined as:
\begin{equation}
    \bm{v}_{t}=\left[\text{LR}^{1,1}_t, \text{LR}^{1,2}_t \cdots \text{LR}^{L,H}_t\right].
\end{equation}

The lookback ratio indicates the model’s focus on context versus its own output during next-token prediction, with a higher ratio suggesting greater emphasis on context.

\section{\gamenospace: Gradient-guided Attention Map Editing}

To mitigate contextual hallucination, we propose a strategic intervention in the attention mechanisms of these LLM models. By editing the attention maps, we aim to enhance the focus on contextual inputs, thereby anchoring the LLMs' outputs more effectively in the provided context and thereby reducing hallucinated contents.

We first conduct a behavioral study to examine the impact of introducing prior biases that augment attention weights towards contextual elements. While this approach yields promising results in summarization tasks, it simultaneously surfaces significant questions about the granularity and specificity of attention manipulation. Overly coarse interventions may inadvertently degrade the generative performance of LLMs.

To address the challenges, we propose \gamenospace, a Gradient-guided Attention Map Editing method, which combines prior bias adjustments with gradient signals obtained from an attention feature-based hallucination classifier to facilitate precise and oriented modifications on attention maps, enabling more effective mitigation of contextual hallucinations.

\subsection{Attention Map Editing with Prior Bias}

We consider linear intervention to the attention map by adding a prior bias ($\bm{b}$) on the original attention map. In the implementation, we add the prior bias to the raw attention scores ($\bm{s}$) before the Softmax normalization step in the attention mechanism. This ensures a valid modified attention map after intervention:
\begin{equation}
        \bm{a} = \text{Softmax}(\bm{s} + \eta\cdot \bm{b}),
\end{equation}
%     \bm{a} = \text{Softmax}(\bm{s} + \eta\cdot \bm{b}),
% \end{align}
where $\eta$ is a hyperparameter that adjusts the intensity of the intervention.

Our design of the prior bias follows two principles: Firstly, the bias should enhance the model's focus on contextual information relative to self-generated content to help mitigate the effects of contextual hallucinations. As Softmax operation is order-preserving, this essentially requires the sum of the bias on the context part should be larger than the generation part, as in \Cref{eq:prior_bias}.
\begin{equation}\label{eq:prior_bias}
    \sum_{i=1}^{N_c} b_{i} > \sum_{j=N_c+1}^{N} b_{j}.
\end{equation}

\begin{figure}[h!]
    \centering
    \includegraphics[width=1.\linewidth]{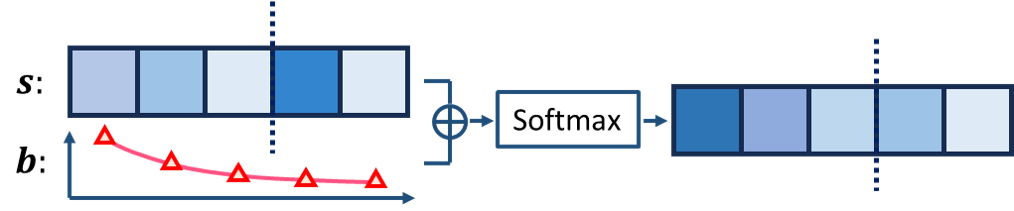}
    \caption{Utilization of the positional-based decay prior attention bias.}
    \label{fig:prior_bias}
\end{figure}

Secondly, it needs to counteract the natural decay of attention that occurs with increasing distance between tokens—a common challenge as discussed in \cite{lost_in_middle, long_context}. Consequently, our bias implementation employs a reverse function and takes in the form: $b_{i} = \frac{1}{i}$, where $b_{i}$ denotes bias of the $i_{\text{th}}$ token when the LLM generates the current token. This function amplifies the model's attention to more distant tokens, effectively counteracting the typical attention decay observed in models like Transformers. A simple illustration of the naive attention editing is shown in \Cref{fig:prior_bias}.
\vspace{-5pt}
\vspace{-5pt}

\paragraph{Empirical Analysis.} 
We apply the naive attention editing uniformly across each attention head in the Llama2-7b model \cite{llama2}. The experiments assess the impact on ROUGE \cite{rouge} scores, testing the summarization capabilities on the XSum dataset. Details of this experimental setup are provided in the Appendix, with results summarized in \Cref{table:exp_prelim}. 

Incorporating a prior attention bias can indeed encourage the model to generate more context-aware outputs, reflected in an increase of the score. However, as $\eta$ increases, the large bias gradually disrupts the original behavior of attention heads and thus leads to dramatic model degradation. 
While the results demonstrate the effectiveness of the prior bias, they also highlight the need for precise attention editing. Specifically, two critical questions arise:
\vspace{5pt}
\begin{itemize}[nosep,leftmargin=10pt]
    \item {\bf Q1}: \emph{when should we apply attention editing?}
    \item {\bf Q2}: \emph{where should we perform attention editing?}
\end{itemize}
\vspace{5pt}
Addressing {\bf Q1} relies on a robust method for detecting contextual hallucination. Given that hallucinations are rare and abnormal events, intervention is necessary only when hallucinated contents are detected. This targeted approach prevents arbitrary bias application, which could otherwise alter the model's desired behavior.

Regarding {\bf Q2}, it is important to recognize that different attention heads exhibit diverse functional focuses \cite{six_head}—some prioritize contextual coherence, while others emphasize content generation. Applying a bias without understanding these distinctions can significantly disrupt their intrinsic behaviors. Therefore, precise identification and selective editing of attention heads are crucial, in determining whether to enhance their focus on contextual coherence or content generation.

\begin{table}[ht]
\centering
    \small
\setlength{\tabcolsep}{2.8mm}{
    % \begin{tabular}{c|c|c|c}
    \begin{tabular}{c|ccc}
    \toprule
    Intensity $\eta$ & ROUGE 1 & ROUGE 2 & ROUGE L \\ \midrule
    0 & 0.2396 & 0.0669 & 0.1682 \\ \midrule
    1 & 0.2422 & 0.0711 & 0.1697 \\ \midrule
    2 & 0.2279 & 0.0634 & 0.1624 \\ \midrule
    5 & 0.0832 & 0.0143 & 0.0624 \\
    \bottomrule
    \end{tabular}
}
\caption{ROUGE score on XSum dataset for Llama2-7b by applying prior attention bias with different edit intensity $\eta$.}
\label{table:exp_prelim}
\end{table}

\begin{figure*}[ht]
    \centering
    \includegraphics[width=0.95\linewidth]{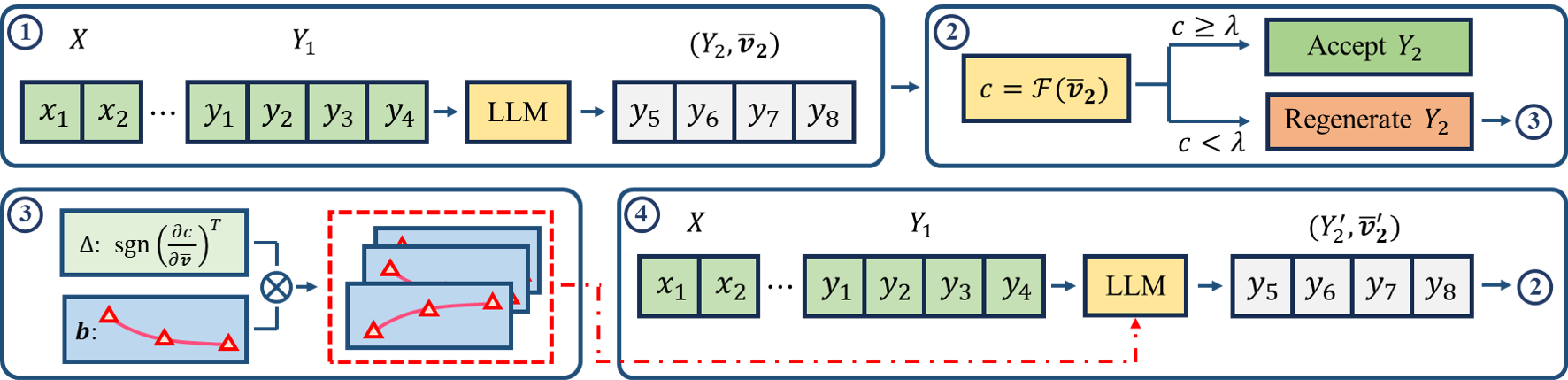}
    \caption{Illustrated example on the generation of one chunk of output in \gamenospace. \textbf{Step \ding{172}}: the LLM predicts the next chunk ($Y_{2}$) and calculates the chunk attention feature ($\bm{\bar{v}_{2}}$) without any attention editing. \textbf{Step \ding{173}}: the classifier ($\mathcal{F}$) predicts the hallucination score ($c$) for the generated chunk with the corresponding feature. If the score exceeds a predefined threshold, the chunk will be accepted. Otherwise, attention editing will be applied to regenerate the chunk. \textbf{Step \ding{174}}: the attention edit signal for each head is computed with the prior bias and the edit direction $\bm{\Delta}$ derived from the gradient of the score. \textbf{Step \ding{175}}: a new chunk is generated with the calculated attention editing signal and re-evaluated with the classifier. If no qualified chunk is accepted with number of regeneration attempts, the chunk with the highest score during the generation process will be accepted.}
    \label{fig:guided_attention-editing}
\end{figure*}

\vspace{-10pt}
\subsection{Gradient-guided Editing}

\game introduces two advanced techniques addressing the issues previously identified. Initially, \game employs a hallucination classifier that utilizes attention features as input to compute a hallucination score. If the generation's score fails to meet a predefined threshold ($\lambda$), it is indicative of hallucination, thereby necessitating attention editing.

Moreover, the classifier not only detects hallucination but also provides gradient information to inform the application of prior biases across different attention heads. This gradient information, termed ``edit direction'' ($\bm{\Delta}$), is a signed binary vector that indicates whether an attention head should increase its focus on the context to elevate the hallucination score, thereby optimizing attention allocation in response to detected hallucinations.

In practice, \game processes outputs in equally sized chunks. An illustrated depiction of the process for generating one chunk is shown in \Cref{fig:guided_attention-editing}. The details of training the classifier and deriving the edit direction are elaborated as follows. To ensure precise and effective modification of the attention maps, and to account for the varied roles of different attention heads, we utilize ``edit direction ($\bm{\Delta}$)'' derived from the gradient information from the classifier. This direction informs how the prior attention bias should be applied to each attention head, optimizing the attention allocation in response to the detected hallucinations. An illustrative overview is depicted in \Cref{fig:guided_attention-editing}.

\subsubsection{Training the Classifier}

\begin{figure}[h!]
    \centering
    \includegraphics[width=0.95\linewidth]{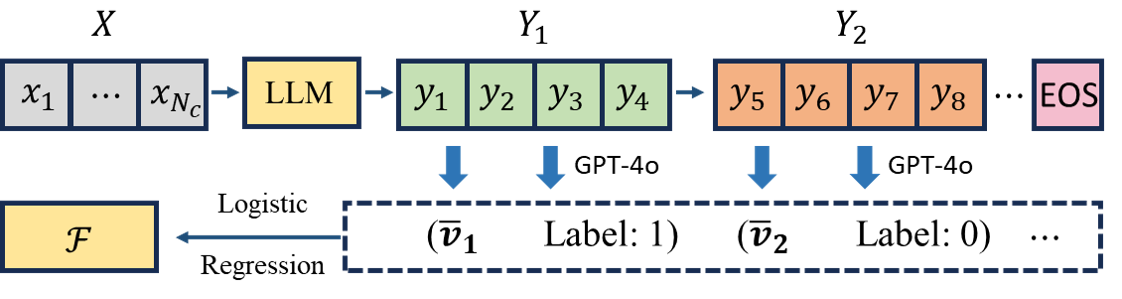}
    \caption{The training data construction and training process of lookback lens.}
    \label{fig:lb_train}
\end{figure}

We train a linear classifier based on the lookback ratio feature to model contextual hallucination. Notably, contextual hallucinated contents usually constitute only a portion of the entire generated text, with the remainder being accurate and relevant. Therefore, it is crucial to model hallucination with greater precision, to accurately capture the correlation between problematic attention features and the corresponding hallucination outputs. To train the classifier, the Llama2-7b model is prompted to generate summaries from a subset of 1,000 articles sampled from the Daily Mail CNN dataset \cite{d_cnn}. The generated outputs are segmented into fixed-size chunks. Each chunk is then annotated by GPT-4o, which assigns binary labels indicating the presence or absence of hallucination, as depicted in \Cref{fig:lb_train}. The attention feature of these chunks, along with their corresponding labels, are used as the training data for the classifier. 

% \vspace{-10pt}
\vspace{-10pt}
% Logistic regression is employed to train the classifier.

\subsubsection{Deriving the Edit Direction $\Delta$.}

\begin{figure}[h!]
    \centering
    \includegraphics[width=1\linewidth]{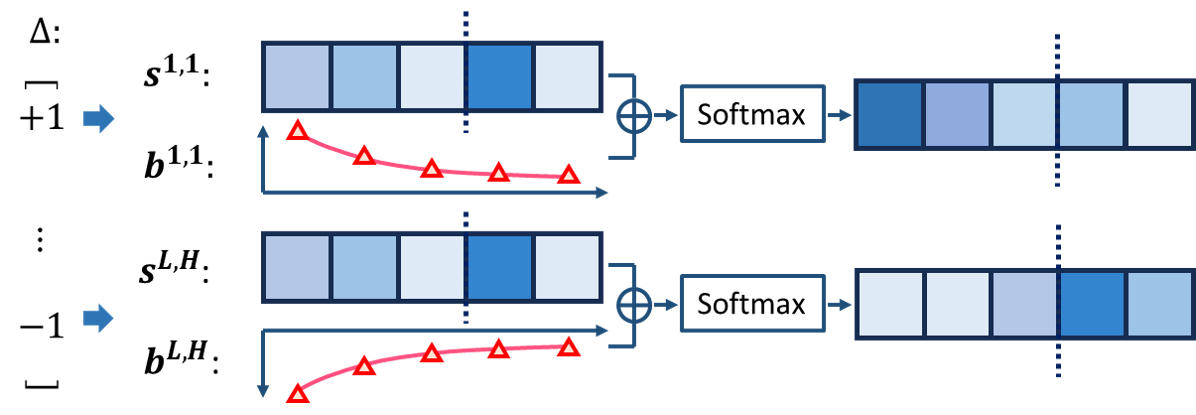}
    \caption{Illustrated example for the combination of prior attention bias and edit direction to perform attention editing.}
    \label{fig:edit_direction}
\end{figure}

Given a detected hallucinated chunk ($Y$) with its corresponding feature $\bar{\bm{v}}$ and the computed score $c$, the edit direction for regenerating this chunk is defined as:
\begin{equation}
    \bm{\Delta} = \text{sgn}(\left[\frac{\partial c}{\partial \bm{\bar{v}}}\right]^{T}),
\end{equation}
where
\begin{equation}
    \text{sgn}(x) = \begin{cases} 
    1 & \text{if } x \geq \epsilon \\
    0 & \text{if } -\epsilon < x < \epsilon \\
    -1 & \text{if } x \leq -\epsilon,
    \end{cases}
\end{equation}
with $\epsilon$ as a pre-defined threshold parameter. During regeneration, the prior bias is multiplied by the edit direction and then added to the original attention, as shown in \Cref{fig:edit_direction}.
\begin{equation}
    \bm{a} = \text{Softmax}(\bm{s} + \eta\cdot  \Delta \cdot \bm{b})
\end{equation}

\vspace{-10pt}
\vspace{-5pt}

\subsection{Interpretation of Edit Direction $\Delta$}
The derivation of $\bm{\Delta}$ incorporates three key considerations to effectively guide attention map editing.

\paragraph{(1) Interpretation of the Gradient.} The gradient term, $\left[\frac{\partial c}{\partial \bm{\bar{v}}}\right]^{T}$, quantifies how modifications to the average lookback ratio of attention features influence the hallucination score. A higher lookback ratio, which indicates a greater focus on contextual information, is generally associated with reduced hallucination. The gradient thus reflects the necessary change in attention focus to mitigate hallucination effects. However, since this gradient is averaged across all tokens within a chunk, direct application in editing is impractical. Instead, we utilize the sign of the gradient (sgn) to determine the general direction for modifying the attention, thereby guiding the regeneration process in a binary manner—either increasing or decreasing focus based on the context.

\paragraph{(2) Effect of Sign Function.} Utilizing the sign function simplifies the gradient information to a directional indicator that instructs whether to enhance or reduce the attention focus on specific elements of the context. This approach avoids the complexities and potential overfitting that might arise from using the precise gradient values, providing a robust mechanism for attention modification.

\paragraph{(3) The Role of Gradient Threshold.} To further refine our approach, we introduce a gradient threshold, $\epsilon$, which serves to filter out attention heads with relatively minor gradients. This selection criterion ensures that only those heads with substantial discrepancies in attention allocation—indicating a strong need for adjustment—are edited. Attention heads with gradients below this threshold are considered adequately aligned and are not subjected to modification. This selective editing helps maintain the model’s overall stability and prevents unnecessary adjustments that could disrupt the model's performance.

\begin{table*}[ht]
\centering
    \small
\setlength{\tabcolsep}{2.6mm}{
    \begin{tabular}{l|cccc}
    \toprule
    \textbf{Models} & \textbf{CNN/Daily Mail} & \textbf{XSum} & \textbf{NQ-Open} & \textbf{Trivial QA} \\ \midrule
    Llama2-base \cite{llama2} & 0.214 & 0.490 & 0.712 & 0.838 \\ \midrule
    Llama2-\game (ours) & {\bf 0.232} & {\bf 0.590} & {\bf 0.732} & {\bf 0.868} \\ \midrule
    Phi3-base \cite{phi3} & 0.203 & 0.504 & 0.690 & 0.803 \\ \midrule
    Phi3-\game (ours) & 0.225 & 0.523 & 0.720 & 0.823 \\
    \bottomrule
    \end{tabular}
}
\caption{Results on summarization and openbook QA for baseline Llama2-7b, Phi3-mini and their corresponding counterpart with the proposed \gamenospace.}
\label{table:exp_1}
\end{table*}

\section{Experiments}

We evaluate the proposed method, \gamenospace, on summarization task and open-book QA task, with two different open source models: Llama2-7B \cite{llama2} and Phi3-mini \cite{phi3}. The detailed configuration of datasets and evaluation metrics for contextual hallucination are summarized below.

\subsection{Experimental Setup}
\paragraph{Datasets.}
We adopt Daily CNN mail \cite{d_cnn} and Extreme Sum (XSum) \cite{xsum} for the summarization task. Following the setup in \cite{lb_lens}, we randomly select 1000 data points from the whole datasets for evaluation.

For the openbook QA task, we adopt the constructed Natural Question \cite{nq} dataset in \cite{lb_lens, lost_in_middle} and follow the sample principle to construct a subset of 1000 data points from the Trivial QA (web) \cite{trivial_qa} dataset. Details for preparing the datasets can be found in \Cref{sec:appendix_dataset}.

\paragraph{Evaluation Metrics.}
For the summarization task, we use GPT-4o as a judge to detect whether there is any hallucination in the summarization. The non-hallucination rate (accuracy) is defined as the number of non-hallucinated summarizations over the total number of data points, which is higher the better. For the openbook QA task, we calculate the best span exact match (EM) rate.

\paragraph{Model Configurations} All baseline models utilize greedy search for decoding. For guided attention editing, we train the classifier using Logistic regression as detailed in \Cref{sec:appendix_exp_details}. This trained classifier is shared by all models across experiments. 

\subsection{Main Results}

We demonstrate that \game improves both accuracy and Exact Match (EM) rate compared to the baseline LLMs, as summarized in \Cref{table:exp_1}, which reports the performance across four datasets and two base models. The proposed \game consistently reduces contextual hallucination and enhancing model performance.

Specifically, on the XSum dataset, Llama2-\game achieves a significant accuracy improvement of $10\%$ over Llama2-base, reducing the number of hallucinated generations from 510 to 410 out of a total of 1000 samples. This improvement highlights the effectiveness of \game in addressing hallucination in challenging summarization tasks.

Similarly, on other datasets such as CNN/Daily Mail and NQ-Open, Llama2-\game demonstrates consistent improvements exceeding $2\%$, underscoring the robustness of the proposed method across diverse domains and task types. On the Trivial QA dataset, applying \game increases the EM rate by $3\%$, improving from $0.838$ to $0.868$.

For the Phi3 model, \game achieves comparable effectiveness, with accuracy improvements of $1.2\%$ on CNN/Daily Mail and $1.9\%$ on XSum. On NQ-Open and Trivial QA, \game enhances the EM rate by $2\%$ and $3\%$ respectively, further validating its versatility. Across all datasets, \game consistently elevates the performance of Phi3-base, demonstrating its generalizability across different LLM architectures.

\paragraph{The Necessity of Edit Direction.}
We additionally analyze the importance of utilizing the editing direction when doing attention editing. We compare a Llama2-7b model utilizing edit direction (denoted by Llama2-\gamenospace-w direction), with a Llama2-7b that uniformly applies the prior attention bias (denoted by Llama2-\gamenospace-w/o direction), on XSum and NQ-Open. The corresponding results are shown in \Cref{table:exp_2}.

Consistent with our preliminary findings, the uniform application of prior attention bias does enhance the model's focus on contextual elements, yielding improved performance relative to the baseline. However, the incorporation of edit direction further enhances overall performance. This supports literature indicating that different attention heads contribute variably to the generation process and underscores the critical need for employing edit direction in the modification of attention maps to optimize model efficacy.

\begin{table}[ht]
\centering
    \small
\setlength{\tabcolsep}{1.5mm}{
    % \begin{tabular}{c|c|c}
    \begin{tabular}{l|cc}
    \toprule
    \textbf{Models}& \textbf{XSum} & \textbf{NQ-Open} \\ \midrule
    Llama2-base \cite{llama2}  & 0.490 & 0.712 \\ \midrule
    Llama2-\game with direction & {\bf 0.590} & {\bf 0.732} \\ \midrule
    Llama2-\game without direction & 0.539 & 0.717\\
    \bottomrule
    \end{tabular}
}
\caption{Results on XSum and NQ-Open for Llama2-\game with direction and Llama2-\game without direction.}
\label{table:exp_2}
\end{table}
\vspace{-10pt}

\subsection{Analysis}

\paragraph{Computational Efficiency.}
Authors in \cite{lb_lens} mitigate hallucination by applying random sampling to generate candidate chunks and accept chunks with the highest score produced by the classifier. This method can suffer from two deficiencies. First, it relies on the model's original ability, assuming the models can eventually generate less hallucinated outputs via repeated sampling. Second, the repetitive sampling process for each chunk reduces the efficiency of the method.

We compare the proposed method and the decoding method in \cite{lb_lens} on XSum and NQ-Open, following the setup in their paper. We summarize the results and the averaged cost, denoted by Run-time (in seconds) per sample on XSum as in \Cref{table:exp_3}. The results show that our proposed method outperforms lookback lens guided decoding on XSum and is {\bf 7X} more efficient. For the NQ-Open dataset, lookback lens guided decoding benefits from high temperature in generating diverse outputs for final selection, but at the cost of low efficiency.

\begin{table}[ht]
\centering
    \small
\setlength{\tabcolsep}{1.4mm}{
    % \begin{tabular}{c|c|c|c}
    \begin{tabular}{l|cc|c}
    \toprule
    \textbf{Models}& \textbf{XSum} & \textbf{NQ-Open} & \textbf{Run-time/sample} \\ \midrule
    Llama2-\game & {\bf 0.590} & {0.732} & 2.69s \\ \midrule
    Lookback Lens & 0.583 & {\bf 0.742} & 18.93s\\
    \bottomrule
    \end{tabular}
}
\caption{Comparison between Llama2-\game and lookback lens guided decoding on XSum and NQ-Open.}
\label{table:exp_3}
\end{table}
\vspace{-10pt}

\paragraph{Comparison of different prior bias.}
The design and choice of the prior attention bias also affects the intervention performance. We consider and compare another uniform bias where $b_{i}=1$ for $i \le N_{c}$ and $b_{i}=0$ for $i > N_{c}$. In our experiments, we compared the performance of the Llama2 model with two different variants: Llama2-\gamenospace-decay and Llama2-\gamenospace-uniform, on the XSum dataset. Both Llama2-decay and Llama2-uniform demonstrated improvements over the baseline Llama2 model. Notably, Llama2-\gamenospace-decay outperformed Llama2-\gamenospace-uniform by an additional ${\bf 5.7\%}$, suggesting that a dynamically scaled bias, which accounts for the positional relevance within the context, is more effective in enhancing model performance.

\begin{table}[ht]
\centering
    \small
\setlength{\tabcolsep}{3.3mm}{
    % \begin{tabular}{c|c}
    \begin{tabular}{l|c}
    \toprule
    \textbf{Models}& \textbf{XSum}\\ \midrule
    Llama2-\gamenospace-decay bias & {\bf 0.590} \\ \midrule
    Llama2-\gamenospace-uniform bias & 0.543 \\
    \bottomrule
    \end{tabular}
}
\caption{Results on XSum for Llama2-\gamenospace-decay bias and Llama2-\gamenospace-uniform bias.}
\label{table:exp_ablation_1}
\end{table}
\vspace{-10pt}

\begin{figure}[h!]
    \centering
    \includegraphics[width=1\linewidth]{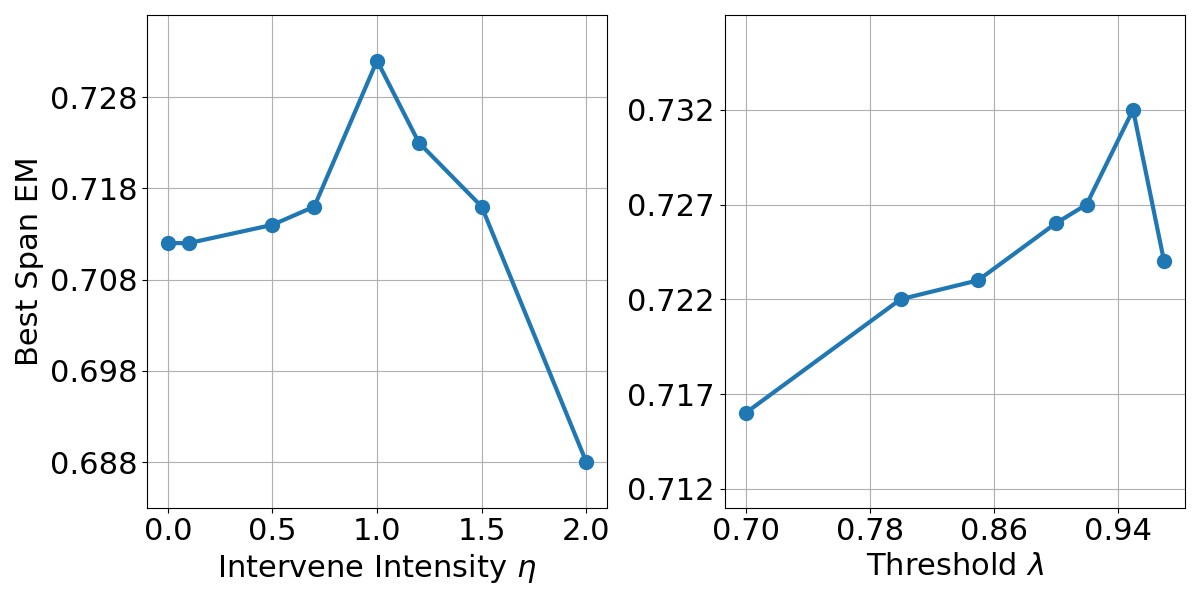}
    \caption{Illustrated results on the impact of different $\eta$ (Left) and different $\lambda$ (Right) on the performance on NQ-Open when applying \gamenospace.}
    \label{fig:ablation}
\end{figure}

\paragraph{Impact of Edit Intensity.}
The hyper-parameter $\eta$ plays a critical role in modulating the extent of intervention during the attention map regeneration process. A value of $\eta$ that is too low may not sufficiently influence the attention map, thereby failing to alter the model's final outputs significantly. Conversely, an excessively high value of $\eta$ can disrupt the intrinsic behavior of the target LLM, leading to compromised generation quality. We systematically evaluated the impact of various settings of $\eta$. using the NQ-Open dataset. Through extensive empirical experiments, we observe that an intensity level above 2 results in nonsensical text generation. Therefore, we have restricted the intensity range to [0, 2]. The results, illustrated in \Cref{fig:ablation} (Left), are consistent with our intuition: moderate increases in $\eta$ enhance the model's focus on relevant context and improve the performance. However, excessively high values result in a dramatic degradation of model behavior.

\vspace{-10pt}

\paragraph{Effect of Editing Threshold.} The threshold $\lambda$ serves as a criterion to decide whether a generated chunk needs regeneration and subsequent attention editing. A higher $\lambda$ imposes stricter criteria for accepting generated chunks and an excessively high threshold can potentially lead to rejection of even well-formed outputs. In such case, \game can inadvertently lead to the unnecessary regeneration of originally accurate chunks, thereby disrupting their quality. We conducted an evaluation of different $\lambda$ values on the NQ-Open dataset using the Llama2-7B model. Empirically, we find that lowering the threshold increasingly allowed more chunks to pass without editing. Notably, setting the threshold below 0.7 resulted in outcomes indistinguishable from the baseline. This empirical observation led us to establish 0.7 as the optimal lower limit in the analysis section. The results, depicted in \Cref{fig:ablation} (Right), indicate that increasing $\lambda$ initially leads to more chunks being flagged for regeneration, which correlates with a performance improvement. However, further increases in $\lambda$ result in a performance decline, as even well-formed chunks are subjected to unnecessary regeneration, leading to outputs that deviate from the model's originally accurate generations.

\section{Related Work}

\paragraph{Advanced Analyses on Attention Mechanisms.}

Attention heads have been found to be closely related to the model behavior and focus on different tasks, from knowledge rescaling to latent reasoning, as summarized in a recent survey \cite{attn_survey}. \citet{attn_dropout} introduces dropout before the feedforward network, aiming to recalibrate attention matrices, focusing more on semantically important tokens to reduce the influence of outlier high-score tokens. \citet{attn_sink} dive into ``attention sinks'' and reveals that certain tokens disproportionately attract attention without adding semantic value. It proposes to recalibrate these attention distributions to enhance LLM reasoning ability. \citet{lb_lens} proposes to detect and mitigate contextual hallucination with a feature, named by ``Lookback Ratio'' derived from the attention map. Differing from these studies, our proposed method, \gamenospace, focuses on actively editing attention maps via gradient-oriented information to control context-aware generation in LLMs. 

\paragraph{Hallucination Mitigation with Representation Editing.} 
Extensive research efforts have been devoted to mitigating various types of hallucinations in LLMs \cite{huang2023survey,tonmoy2024comprehensive,chen2024context,ji2023towards,luo2024hallucination,wang2024developing}. A significant line of works propose to intervene on hidden representation to mitigate hallucination. For instance, Plug and Play \cite{plug_and_play} leverages the gradient of an attribute model to adjust the hidden representations, guiding LLMs toward generating outputs with specific desired attributes. ITI \cite{iti} employs neural probing \cite{neural_probe} to classify attention head outputs, suggesting adjustments to activations during inference to enhance factual correctness. \citet{truthx} integrates an auto-encoder to split the hidden representation into components related to factual and semantic content, thus enabling more precise control. 
\citet{dola} posits that factual information is progressively revealed across decoder layers and introduces a method of contrastive decoding between layers to highlight this dynamic. These methodologies primarily address the mitigation of closed-book hallucination, as defined in \cite{constructing_benchmark_hallucination}. However, \game diverges from these approaches in two key aspects. Firstly, \game specifically targets contextual hallucination, rather than the closed-book scenario. Secondly, it is inspired by the observed correlations between attention mechanisms and contextual hallucination, directly intervening at the level of the attention map to influence output generation.

\section{Conclusion}
We introduce \gamenospace, a novel method designed to perform precise attention map editing to counteract contextual hallucination. This is achieved by incorporating meticulously designed prior bias and gradient information derived from hallucination classifiers. \game has been rigorously tested across two open-source LLMs on four distinct datasets, effectively demonstrating its capability to mitigate contextual hallucination. Future research will aim to further enhance the efficiency and explore its applicability to a broader range of tasks.

% \newpage
\section{Limitations}
While the proposed guided attention editing method demonstrates significant improvements in reducing contextual hallucination, this work still presents limitations that pave the way for promising future research. First, the prior attention bias, though effective, is currently heuristic-based and could potentially be optimized through learning from a small dataset to more accurately intervene in the model's behavior. Second, our method adheres to a ``detect then mitigate'' paradigm, where hallucinations are identified and rectified post-generation in discrete chunks. However, the propensity of the model to generate hallucinated content might be identifiable prior to actual content generation. Early detection and prediction of potential hallucinations represent a compelling direction for future research, which could lead to more proactive strategies in managing and mitigating errors in LLMs.
\bibliography{reference}

\newpage
\newpage
\clearpage

\appendix
\section{Dataset Details}\label{sec:appendix_dataset}
\subsection{CNN/Daily Mail}
CNN/Daily Mail \cite{d_cnn} is originally designed for machine-reading and text understanding, the CNN/Daily Mail dataset consists of news articles and their respective highlights, enabling models to practice summarization tasks. It is used by us in training the classifier and evaluation dataset in summarization task. For the training of the classifier, we follow \cite{lb_lens} to randomly sample 1000 data from the testing set. For summarization task evaluation, we sampled another 1000 samples from the testing set. 

The dataset uses the Apache-2.0 license and can be found at: \url{https://huggingface.co/datasets/abisee/cnn_dailymail}.

\subsection{XSum}
The Extreme Summarization (XSum) \cite{xsum} dataset is tailored for generating single-sentence news summaries, presenting a challenge in capturing the main point of an article with minimal context. In this paper, we randomly sampled 1000 data from the testing dataset for evaluation.

The dataset uses the MIT license and can be found at: \url{https://github.com/EdinburghNLP/XSum}.

\subsection{Natural Questions}
Natural Questions (NQ) \cite{nq} is developed by Google and contains real user questions sourced from Google search, paired with Wikipedia article answers. We follow \cite{lb_lens} to construct its openbook variant (NQ-Open) by randomly sampling 1000 data points from a processed dataset \cite{lost_in_middle}, which can be found at\url{https://github.com/nelson-liu/lost-in-the-middle}. Specifically, the input contexts are created by concatenating three different source documents, where the first and third documents are irrelevant and the second document contains the relevant information.

The original NQ dataset uses the Apache-2.0 license and can be found at: \url{https://github.com/google-research-datasets/natural-questions}.

\begin{table*}[h!]
\centering
\small
\begin{tabular}{p{\dimexpr \linewidth-2\tabcolsep}}
\toprule
\begin{minipage}[t]{\linewidth}
\texttt{\textbf{CNN/Daily Mail:} Generate a summary based on the information in the document.\\
\textbf{XSum:} Generate a summary comprising of 1 sentence for the given article.\\
\textbf{NQ-Open:} Answer the question based on the information in the document. Explain your reasoning in the document step-by-step before providing the final answer.\\
\textbf{Trivial QA:} Answer the question based on the information in the document. Explain your reasoning in the document step-by-step before providing the final answer.\\
}
\end{minipage} \\
\bottomrule
\end{tabular}
\caption{System prompts for different datasets used by Llama2-7b and Phi3-mini.}
\label{tab:dataset_prompt}
\end{table*}

\begin{table*}[h!]
\centering
\small
\begin{tabular}{p{\dimexpr \linewidth-2\tabcolsep}}
\toprule
\begin{minipage}[t]{\linewidth}
\texttt{You will be provided with a document and a proposed summary. Your task is to determine if the proposed summary can be directly inferred from the document. If the summary contains any information not found in the document, it is considered false. Even if the summary is different from a ground truth summary, it might still be true, as long as it doesn't contain false information.\\
For each proposed summary, explain why it is true or false based on the information from the document. Focus only on the original document's content, disregarding any external context.\\
After your explanation, give your final conclusion as \red{Conclusion: True} if the proposed summary is completely accurate based on the document, or \red{Conclusion: False} if it contains any incorrect or unsupported information. If your conclusion is 'False', identify the exact phrases or name entities from the summary that is incorrect by stating \red{Problematic Spans: [the inaccurate text spans from the summary, in Python list of strings format]}.\\
\\
\#Document\#: \{document\}\\
\\
\#Ground Truth Summary\#: \{ground\_truth\_summary\}\\
\\
\#Proposed Summary\#: \{response\}\\
\\
Write your explanation first, and then give your final conclusion as \red{Conclusion: True} if the proposed summary is completely accurate based on the document, or \red{Conclusion: False} if it contains any incorrect or unsupported information. Add \red{Problematic Spans: [the exact inaccurate text spans from the summary, in a list of strings]} if your conclusion is 'False'.}
\end{minipage} \\
\bottomrule
\end{tabular}
\caption{Prompts for GPT-4o to annotate the span-level hallucinations for given generation from LLMs on summarization tasks (CNN/Daily Mail and XSum).}
\label{tab:gpt4-prompt-sum}
\end{table*}

\subsection{Trivial QA}
Trivial QA \cite{trivial_qa} is a collection of trivia question-answer pairs with supporting documents from Wikipedia (trivial QA-wiki) or from web search results (trivial QA-web). We randomly sampled 1000 data points from the trivial QA-web. For each data point, we select the source document with the highest score as the context. During the construction, we remove instances whose context length is larger than the base models' context window.

The original dataset uses the Apache-2.0 license and can be found at: \url{https://github.com/mandarjoshi90/triviaqa}.

\section{Models and License}
We utilize two open-source models in this paper, both are adopted from their Huggingface Transformer implementation. Llama2-7b (model ID: meta-llama/Llama-2-7b-hf) \url{https://huggingface.co/meta-llama/Llama-2-7b} is a 7B parameter, instructional finetuned LLM by Meta, under Llama 2 Community License Agreement. Phi3-mini (Model ID: Microsoft/Phi-3-mini-4k-instruct) \url{https://huggingface.co/microsoft/Phi-3-mini-4k-instruct} is a 2B parameter, instructional finetuned LLM by Microsoft.

\section{Detailed Experimental Setup}\label{sec:appendix_exp_details}
\subsection{LLM Setting}
We adopt greedy search in baseline models and their variants that are applied with \gamenospace. The number of maximum new tokens is set to 256. When using \gamenospace, the LLM generates output in chunks and the chunk size is set to 8.
\subsection{Classifier Setting} 
We utilize 1000 samples from the CNN/Daily Mail dataset to train the classifier. The Llama2-7b model is prompt (see the prompt in \Cref{sec:appendix_prompts}) to generate summarization on these examples. The outputs are divided into chunks with a size of 8 tokens each. We utilize GPT-4o as an annotator as described in \Cref{sec:appendix_gpt_4o} to create labels for each chunk. The labels as well as the Lookback Ratio of each chunk are used as the training data for a linear logistic regression classifier with scikit-learn\footnote{\url{https://scikit-learn.org/1.5/modules/generated/sklearn.linear_model.LogisticRegression.html}}.

Due to the rarity of hallucination phenomena, the training data is highly imbalanced. To address this, we additionally drew 200 samples to determine the default threshold. Setting this threshold at 0.9 resulted in the highest test AUROC. The default $\epsilon$ in the $\text{sgn}(\cdot)$ function is set to be 1e-4.

\subsection{Prompts for Different Datasets}\label{sec:appendix_prompts}
We use the same system prompts for both Llama2-7b and Phi3-mini as shown in \Cref{tab:dataset_prompt}.

\section{GPT-4o Annotation Prompts}\label{sec:appendix_gpt_4o}
We utilize GPT-4o (model version: gpt-4o-2024-05-13) to annotate hallucinated spans in the generation of LLMs to prepare training data for the classifier. It is also utilized to judge whether contextual hallucination happens in our evaluation on summarization datasets. For both tasks, we adopt the same prompt from \cite{lb_lens}, detailed as in \Cref{tab:gpt4-prompt-sum}:

\section{Code Implementation and Computation Resources}
The code of the paper is developed based on Huggingface Transformer (4.42.0) \url{https://github.com/huggingface/transformers} and part of the code is adopted from Lookback Lens \url{https://github.com/voidism/Lookback-Lens}. The application of \game requires no training or finetuning of LLMs. All experiments can be run on a single Nvidia Ampere 100 (80GB) GPU. The average inference time per sample in the XSum dataset is around 3 seconds. For other datasets, the time may vary based on the input context length.

% \section{Discussion and Future Works}
% While the proposed guided attention editing method demonstrates significant improvements in reducing contextual hallucination, this work still presents limitations that pave the way for promising future research. Firstly, the prior attention bias, though effective, is currently heuristic-based and could potentially be optimized through learning from a small dataset to more accurately intervene in the model's behavior. Secondly, our method adheres to a ``detect then mitigate" paradigm, where hallucinations are identified and rectified post-generation in discrete chunks. However, the propensity of the model to generate hallucinated content might be identifiable prior to actual content generation. Early detection and prediction of potential hallucinations represent a compelling direction for future research, which could lead to more proactive strategies in managing and mitigating errors in LLMs.

\end{document}